\documentclass[letterpaper, 10 pt, conference]{ieeeconf}
\usepackage{bm}
\usepackage{amssymb}
\usepackage{mathtools}
\usepackage{nicefrac}  
\usepackage[noend]{algpseudocode}
\usepackage{algorithm}
\usepackage{setspace}
\usepackage{booktabs}
\usepackage{multicol}
\usepackage{multirow}
\usepackage{tabularx}
\usepackage{etoolbox}
\newrobustcmd{\B}{\bfseries}  
\usepackage{siunitx}  
\usepackage{diagbox}
\usepackage{pifont}  
\usepackage{todo}  
\usepackage{color}
\usepackage{xcolor}
\definecolor{red}{RGB}{200,80,80}
\definecolor{green}{RGB}{80,200,80}
\definecolor{blue}{RGB}{80,80,200}
\definecolor{orange}{RGB}{243,147,82}
\definecolor{purple}{RGB}{150,100,250}
\definecolor{yellow}{RGB}{231,198,100}
\definecolor{gray}{RGB}{98,95,75}
\usepackage{lipsum}
\usepackage{times}
\usepackage{graphicx}
\usepackage[caption=false,font=footnotesize]{subfig}
\usepackage{wrapfig}  
\usepackage{tikz}
\usetikzlibrary{positioning,shapes,bayesnet,calc,backgrounds}

\makeatletter
\newif\if@showgrid@grid
\newif\if@showgrid@left
\newif\if@showgrid@right
\newif\if@showgrid@below
\newif\if@showgrid@above
\tikzset{%
    every show grid/.style={},
    show grid/.style={execute at end picture={\@showgrid{grid=true,#1}}},%
    show grid/.default={true},
    show grid/.cd,
    labels/.style={font={\sffamily\small},help lines},
    xlabels/.style={},
    ylabels/.style={},
    keep bb/.code={\useasboundingbox (current bounding box.south west) rectangle (current bounding box.north west);},
    true/.style={left,below},
    false/.style={left=false,right=false,above=false,below=false,grid=false},
    none/.style={left=false,right=false,above=false,below=false},
    all/.style={left=true,right=true,above=true,below=true},
    grid/.is if=@showgrid@grid,
    left/.is if=@showgrid@left,
    right/.is if=@showgrid@right,
    below/.is if=@showgrid@below,
    above/.is if=@showgrid@above,
    false,
}
\def\@showgrid#1{%
    \begin{scope}[every show grid,show grid/.cd,#1]
    \if@showgrid@grid
    \begin{pgfonlayer}{background}
    \draw [help lines]
        (current bounding box.south west) grid
        (current bounding box.north east);
    \pgfpointxy{1}{1}%
    \edef\xs{\the\pgf@x}%
    \edef\ys{\the\pgf@y}%
    \pgfpointanchor{current bounding box}{south west}
    \edef\xa{\the\pgf@x}%
    \edef\ya{\the\pgf@y}%
    \pgfpointanchor{current bounding box}{north east}
    \edef\xb{\the\pgf@x}%
    \edef\yb{\the\pgf@y}%
    \pgfmathtruncatemacro\xbeg{ceil(\xa/\xs)}
    \pgfmathtruncatemacro\xend{floor(\xb/\xs)}
    \if@showgrid@below
    \foreach \X in {\xbeg,...,\xend} {
        \node [below,show grid/labels,show grid/xlabels] at (\X,\ya) {\X};
    }
    \fi
    \if@showgrid@above
    \foreach \X in {\xbeg,...,\xend} {
        \node [above,show grid/labels,show grid/xlabels] at (\X,\yb) {\X};
    }
    \fi
    \pgfmathtruncatemacro\ybeg{ceil(\ya/\ys)}
    \pgfmathtruncatemacro\yend{floor(\yb/\ys)}
    \if@showgrid@left
    \foreach \Y in {\ybeg,...,\yend} {
        \node [left,show grid/labels,show grid/ylabels] at (\xa,\Y) {\Y};
    }
    \fi
    \if@showgrid@right
    \foreach \Y in {\ybeg,...,\yend} {
        \node [right,show grid/labels,show grid/ylabels] at (\xb,\Y) {\Y};
    }
    \fi
    \end{pgfonlayer}
    \fi
    \end{scope}
}
\makeatother
\tikzset{every show grid/.style={show grid/keep bb}}%

\makeatletter
\newcommand*{\T}{{\mathpalette\@transpose{}} }
\newcommand*{\@transpose}[2]{\raisebox{\depth}{$\m@th#1\intercal$}}
\makeatother
\DeclarePairedDelimiterX{\norm}[1]{\lVert}{\rVert_{2}}{#1}
\DeclareMathOperator*{\argmax}{arg\,max}

\makeatletter
\newcommand\thefontsize{The current font size is: \f@size pt}
\makeatother

\makeatletter
\let\NAT@parse\undefined
\makeatother
\usepackage[numbers,sort&compress,square,comma]{natbib}  
\usepackage{hyperref}
\hypersetup{
  colorlinks=true,
  urlcolor=blue,
}
\usepackage{cleveref}  
\crefname{lemma}{Lemma}{Lemmas}
\crefname{proposition}{Proposition}{Propositions}
\crefname{definition}{Definition}{Definitions}
\crefname{theorem}{Theorem}{Theorems}
\crefname{conjecture}{Conjecture}{Conjectures}
\crefname{corollary}{Corollary}{Corollaries}
\crefname{section}{Section}{Sections}
\crefname{appendix}{Appendix}{Appendices}
\crefname{figure}{Fig.}{Figs.}
\crefname{equation}{Eq.}{Eqs.}
\crefname{table}{Table}{Tables}
\crefname{algocf}{Algorithm}{Algorithms}
\usepackage{balance}  
\apptocmd{\sloppy}{\hbadness 10000\relax}{}{}

\IEEEoverridecommandlockouts  
\title{\LARGE \bf Informative Planning in the Presence of Outliers}
\author{Weizhe Chen and Lantao Liu
\thanks{The authors are with the Luddy School of Informatics, Computing, and Engineering, Indiana University, Bloomington, IN, 47408, USA. \{chenweiz, lantao\}@iu.edu.}
}%

\begin{document} 
\maketitle
\thispagestyle{empty}
\pagestyle{empty}
\begin{abstract} 
Informative planning seeks a sequence of actions that guide the robot to collect the most informative data to build a large-scale environmental model or learn a dynamical system. Existing work in informative planning mainly focuses on proposing new planners and applying them to various robotic applications such as environmental monitoring, autonomous exploration, and system identification. The informative planners optimize an objective given by a probabilistic model, e.g., Gaussian process regression (GPR). In practice, the ubiquitous sensing outliers can easily affect the model, resulting in a misleading objective. A straightforward solution is to filter out the outliers in the sensing data stream using an off-the-shelf outlier detector. However, informative samples are also scarce by definition so they might be falsely filtered out. In this paper, we propose a method to enable the robot to re-visit the locations where outliers were sampled besides optimizing the informative planning objective. The robot can collect more samples in the vicinity of outliers and update the outlier detector to reduce the number of false alarms. We achieve this by designing a new objective for the Pareto Monte Carlo tree search (MCTS). We demonstrate that the proposed framework performs better than applying an outlier detector naively.
\end{abstract}
\section{Introduction}
\label{sec:introduction}%
Intelligent robots need to sense the environment with onboard sensors and use the collected data to understand their surroundings before performing other subsequent tasks. \emph{Informative planning}~\cite{singh2007efficient} seeks a sequence of actions that allow the robot to obtain the most informative data, i.e., the data that contributes the most to learning the environment model while minimizing the cost of traveling and sampling. Typically, the level of informativeness for sampling data from a certain location is quantified by the reduction of predictive uncertainty in a probabilistic model. For example, when dealing with spatial phenomena, a common choice for the probabilistic model is \emph{Gaussian process regression} (GPR), and \emph{mutual information} or \emph{variance/entropy reduction}  are commonly used to quantify the information contained in the data~\cite{cao2013multi,ma2018data}.

\begin{figure}[tp]%
  \centering%
  \subfloat[Chlorophyll level over time]{%
    \resizebox{0.5\linewidth}{!}{
      \begin{tikzpicture}%
        \node at(0.0,0.0){\includegraphics[width=\linewidth,height=0.6\linewidth]{%
        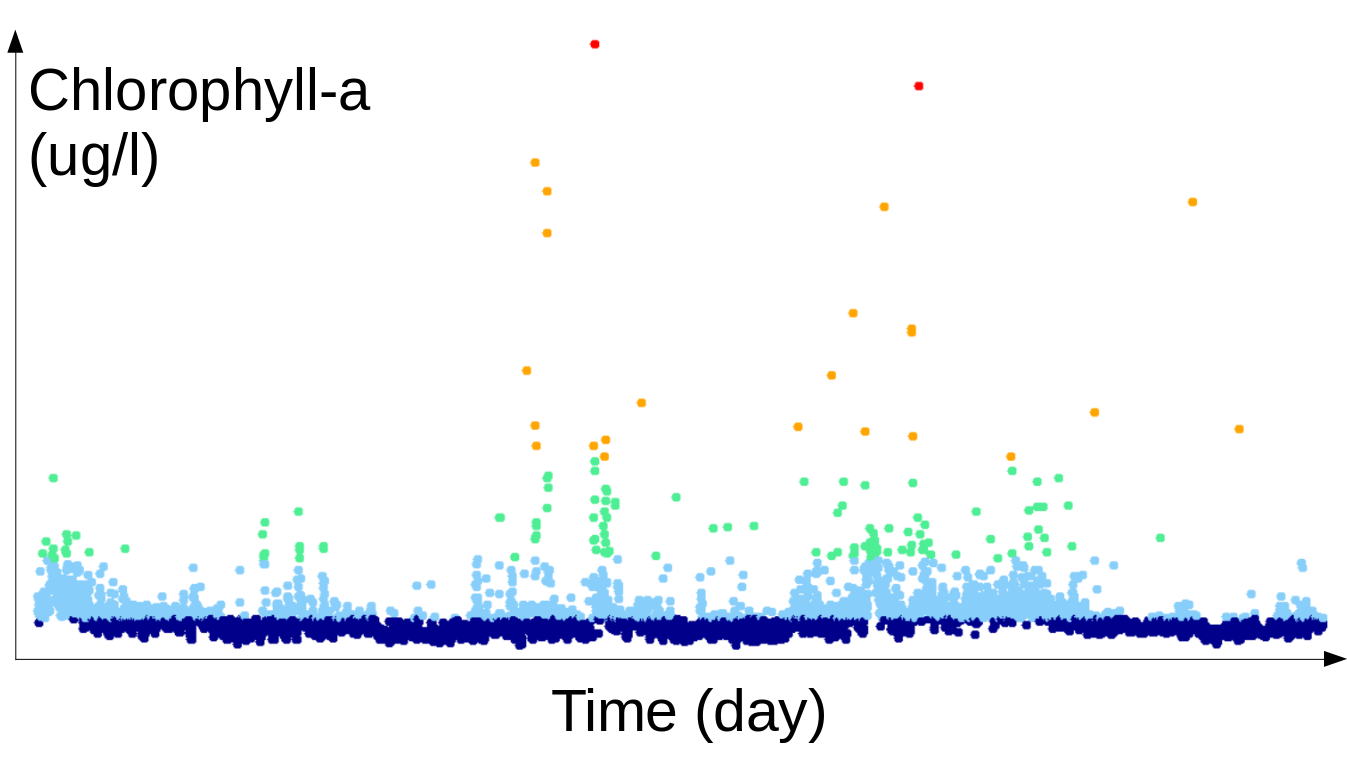}};%
      \end{tikzpicture}}}%
  \subfloat[Ground-truth model and data]{%
    \resizebox{0.5\linewidth}{!}{
      \begin{tikzpicture}%
        \node at(0.0,0.0){\includegraphics[width=\linewidth,height=0.6\linewidth]{%
        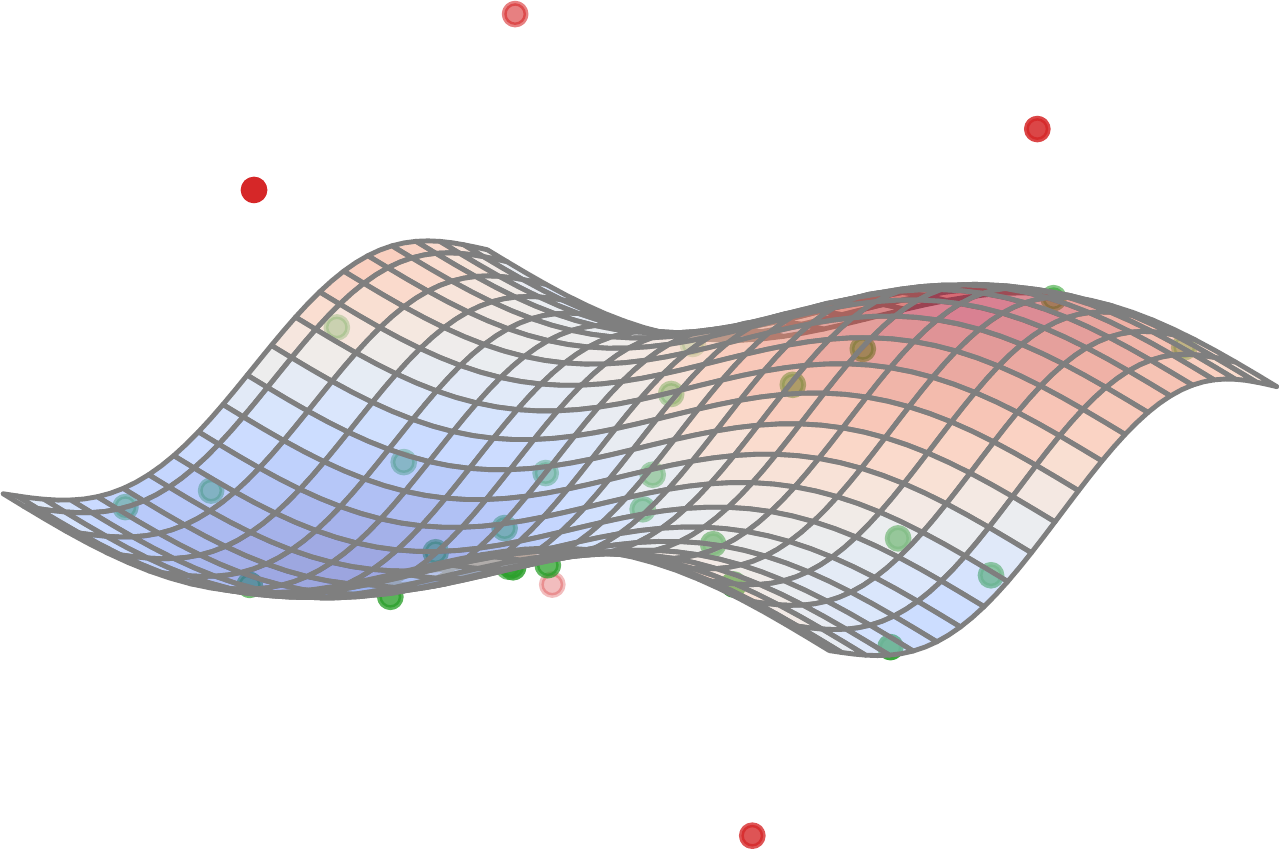}};%
      \end{tikzpicture}}}\\%
  \subfloat[GPR without outliers]{%
    \resizebox{0.5\linewidth}{!}{
      \begin{tikzpicture}%
        \node at(0.0,0.0){\includegraphics[width=\linewidth,height=0.6\linewidth]{%
        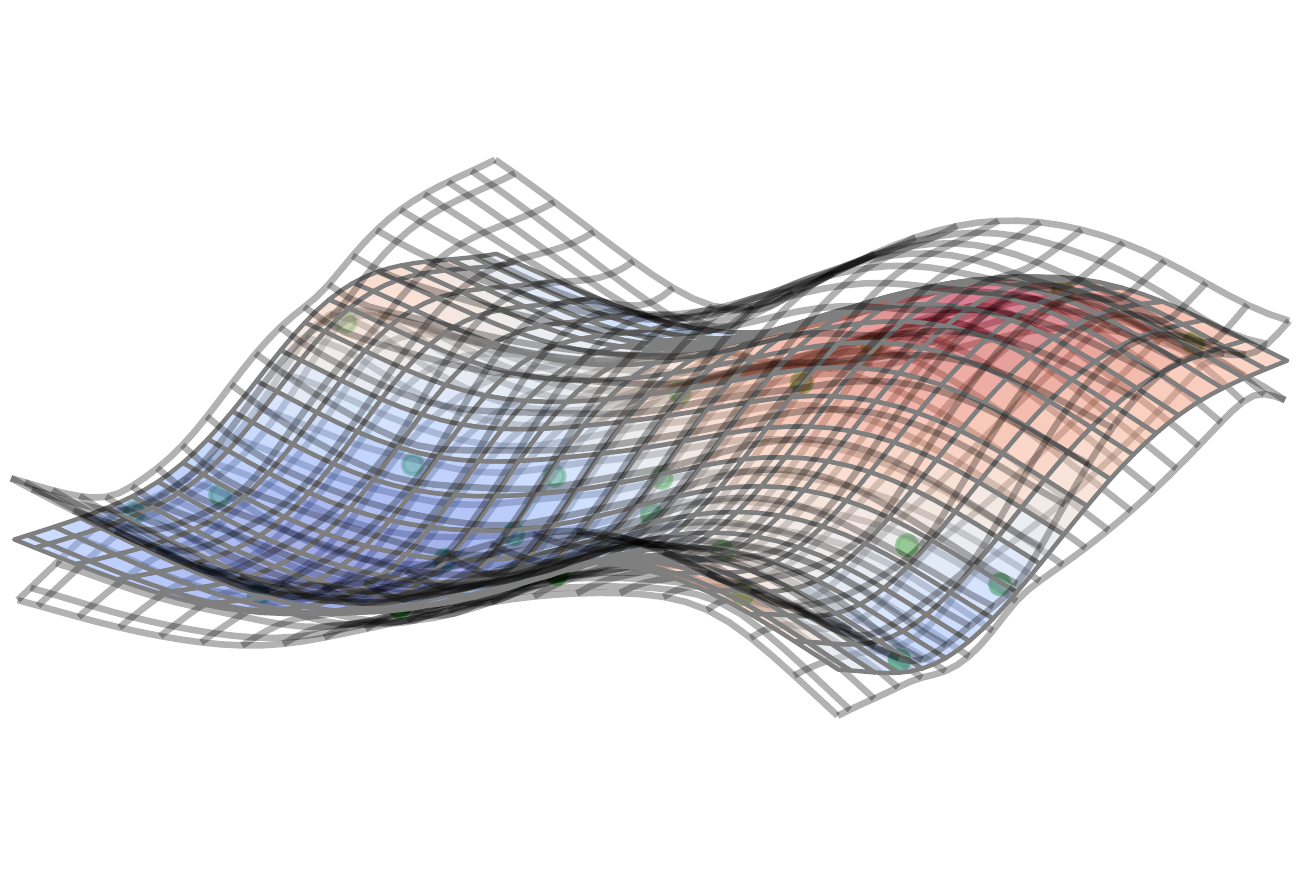}};%
      \end{tikzpicture}}}%
  \subfloat[GPR with outliers]{%
    \resizebox{0.5\linewidth}{!}{
      \begin{tikzpicture}%
        \node at(0.0,0.0){\includegraphics[width=\linewidth,height=0.6\linewidth]{%
        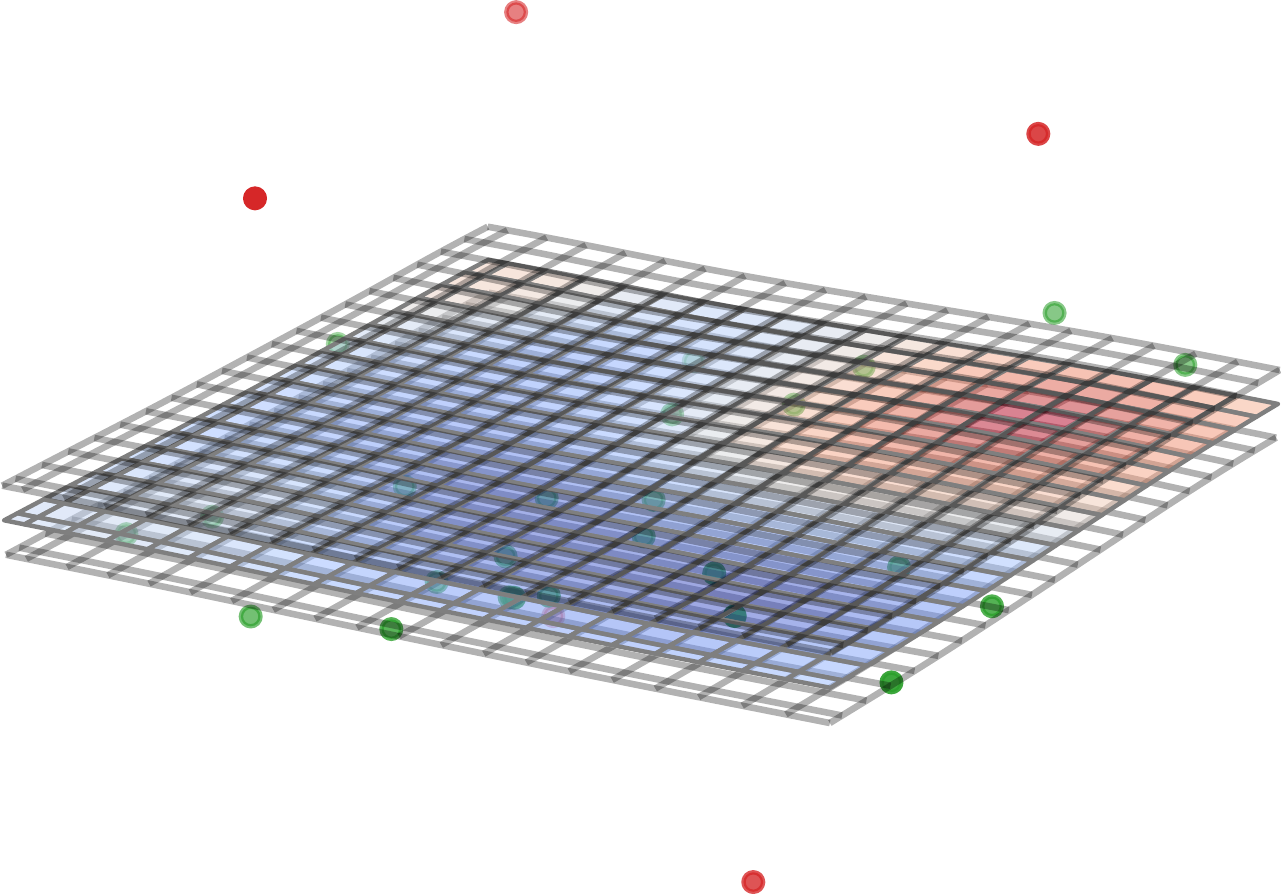}};%
      \end{tikzpicture}}}%
  \caption{\textbf{(a) Chlorophyll-a sensing data}. The orange and red points are outliers~\cite{espinola2017statistical}. \textbf{(b) Ground-truth environmental model}. Green points are normal data points, and red points are outliers. Red and blue colors on the surface indicate high and low values, respectively. \textbf{(c) GPR without outliers} accurately predicts the environment and provides informative confidence bounds (i.e., the upper and lower black grids). \textbf{(d) GPR with outliers} leads to inaccurate ``flat'' prediction and uninformative uncertainty estimates, which further affects the performance of downstream informative planners.}%
  \label{fig:pull_figure}%
\end{figure}

Many informative planners have been proposed during the last two decades~\cite{schmid2020efficient, popovic2017online, best2019dec, hollinger2014sampling, marchant2012bayesian, choudhury2018data}. The majority of the literature assumes the sensing data is the ground-truth value corrupted by Gaussian noise. In practice, however, sensing data typically contains outliers due to various reasons. For instance, when facing specular reflection, LiDAR returns false values of maximum range even though the beams hit obstacles. When multiple exteroceptive sensors are in use, cross-talks among them also cause ``random measurements''~\cite{thrun2002probabilistic}. In outdoor environments, unexpected fast transient obstacles, such as insects, fish, or large dust particles, can also induce inevitable outliers. In addition, sensor wear and tear (e.g., worn probes), among other incomprehensible reasons that cause occasional hardware malfunction, can also introduce outliers. A Gaussian model can be easily distorted by outliers, rendering the ``optimal'' informative sampling decisions computed from such a model sub-optimal. \cref{fig:pull_figure} illustrates a scenario where outliers mislead the modeling and planning performances.

A straightforward idea to mitigate the influence of outliers is to filter them out from the sensing data stream using an off-the-shelf outlier detector. However, this leads to a high false-alarm rate because informative samples and outliers are statistically similar: \emph{they both change the model significantly and are observed less frequently. } Otherwise, the informative samples would not bring much additional information. To reduce the high false-alarm rate, we propose to optimize two objectives simultaneously. In addition to the original information-seeking objective, the other objective encourages re-visiting the locations where outliers were sampled. In this way, the robot can collect more samples in the region where outliers are detected frequently and update the outlier detector to reduce the number of false alarms. We implement this idea by designing a new objective in the recently proposed Pareto Monte Carlo tree search~\cite{chen2019pareto}. We show that the proposed framework performs better than naively applying a state-of-the-art outlier detector.

Although dealing with outliers is essential in practical applications of informative planning algorithms, it has not been well discussed in the literature to the best of our knowledge. The contribution of this paper is to point out the interesting interplay between informative samples and outliers: outliers can also be viewed as ``extremely informative samples''. We developed a practical approach to significantly reduce the number of false alarms due to the interplay.

\section{Related Work}%
\label{sec:literature}%
Informative planning has attracted growing interest, especially in the application of autonomous environmental monitoring~\cite{dunbabin2012robots}. Informative planning extends the optimal sensor placement problem~\cite{krause2008near} by considering the traversal costs and motion constraints of the mobile robots. Starting from the seminal work~\cite{chekuri2005recursive}, various recursive greedy algorithms have been proposed~\cite{singh2007efficient, meliou2007nonmyopic, hollinger2009efficient}. These methods are based on the \emph{submodularity} of the objective function and provide a performance guarantee. The submodularity requirement has been relaxed to \textit{monotonicity}~\cite{binney2013optimizing}. Dynamic programming (DP) based methods do not require the objective function to have these special properties. In~\cite{cao2013multi}, a sequence of informative waypoints is selected via DP by assuming the underlying map to be in a rectangular and sliced shape. This framework was extended to arbitrary continuous space by connecting the informative waypoints via traveling salesman problem solver~\cite{ma2016information}. This framework is lifted to online planning by integrating sparse GPs~\cite{ma2018data}. To develop an efficient planner, \citet{hollinger2014sampling} extend sampling-based motion planners to robotic information gathering algorithms. Recently, this framework has been extended to online variants~\cite{schmid2020efficient, ghaffari2019sampling}. Monte-Carlo tree search (MCTS) based methods are conceptually similar to sampling-based informative planners and have recently garnered great  attention~\cite{arora2018multi,best2019dec,morere2017sequential,chen2019pareto,flaspohler2019information}. Instead of randomly growing the search tree, the MCTS expands tree nodes in a best-first search manner~\cite{browne2012survey}. Since Gaussian processes have been the \textit{de facto} standard for modeling spatiotemporal phenomena in many environmental monitoring applications, Bayesian optimization becomes a natural choice for informative planning~\cite{marchant2012bayesian, bai2016information, ling2016gaussian}. New frameworks for informative planning are constantly emerging with attractive mechanisms such as evolutionary methods~\cite{popovic2017online, Popovic2020informative, hitz2017adaptive} and imitation learning~\cite{choudhury2018data}. Many of the existing works employ GPR as the probabilistic model. Compared to the large body of work in informative planners, investigations on the probabilistic model in the informative planning framework are relatively sparse. Methods using online sparse GPs~\cite{ma2017informative} have been proposed to tackle the computational bottleneck. The mixture of GPs~\cite{luo2018adaptive, ouyang2014multi} has also been applied to capture environmental non-stationarity. Different from the existing efforts, we investigate the effects of sensing outliers on model learning and informative planning. To the best of our knowledge, this is the first time that outliers in informative planning have been discussed.

\section{Problem Formulation}%
\label{sec:problem}
Let $\mathcal{T}\triangleq\{1,2,\dots,T\}$ be the set of decision epochs. At time $t\in\mathcal{T}$, a robot with a fully observed state $\mathbf{s}_{t}\in\mathcal{S}$ collects scalar measurement $y_{t}$ via its sensor and takes an action $a_{t}$, arriving at the next state $\mathbf{s}_{t+1}$ with transition probability $p(\mathbf{s}_{t+1}|\mathbf{s}_{t},a_{t})$. In this work, we assume the transition to be known and deterministic, and focus on inferring the noise-free environment state $f_{\text{env}}$ from its noisy observation collected so far $\mathbf{y}=[y_{1},\dots,y_{N}]^{\T}$. One example is to estimate an elevation map from range sensor measurements. In this case, the elevation map is a function $f_{\text{env}}(\mathbf{x})$ of the input spatial locations $\mathbf{x}\in\mathbb{R}^{D}$. The robot maintains an internal model $f$ of the environment learned from the noisy observations $\mathbf{y}$. The model is updated after taking new observations into account. Ideally, informative planning should use error reduction as the reward signal for the planner. Let $f_{e}(\bm{a})$ be a function that quantifies the modeling error reduction after incorporating the data obtained by taking an action sequence $\bm{a}$, and $f_{c}(\bm{a})$ be a function that returns the cost of executing $\bm{a}$. Informative planning solves the following optimization problem given an available budget $B$,
\begin{equation}
    \bm{a}^{\star}=\argmax_{\bm{a}\in\bm{\mathcal{A}}}f_{e}(\bm{a}), ~s.t.~f_{c}(\bm{a})\leq{B},
\end{equation}
where $\mathcal{A}$ is the space of all possible action sequences. However, the ground-truth environment state $f_{\text{env}}$ is unknown, hindering the computation of error reduction. Informative planning bypasses this problem by utilizing a surrogate reward function $f_{i}(\mathbf{a})$ given by a probabilistic model instead of $f_{e}(\mathbf{a})$. Typically, $f_{i}(\bm{a})$ quantifies the informativeness of the data obtained by taking actions $\bm{a}$ by measuring the reduction in predictive uncertainty of the probabilistic model.

\section{Method}
\label{sec:method}%
\begin{figure}[tp]
  \centering
  \includegraphics[width=\linewidth]{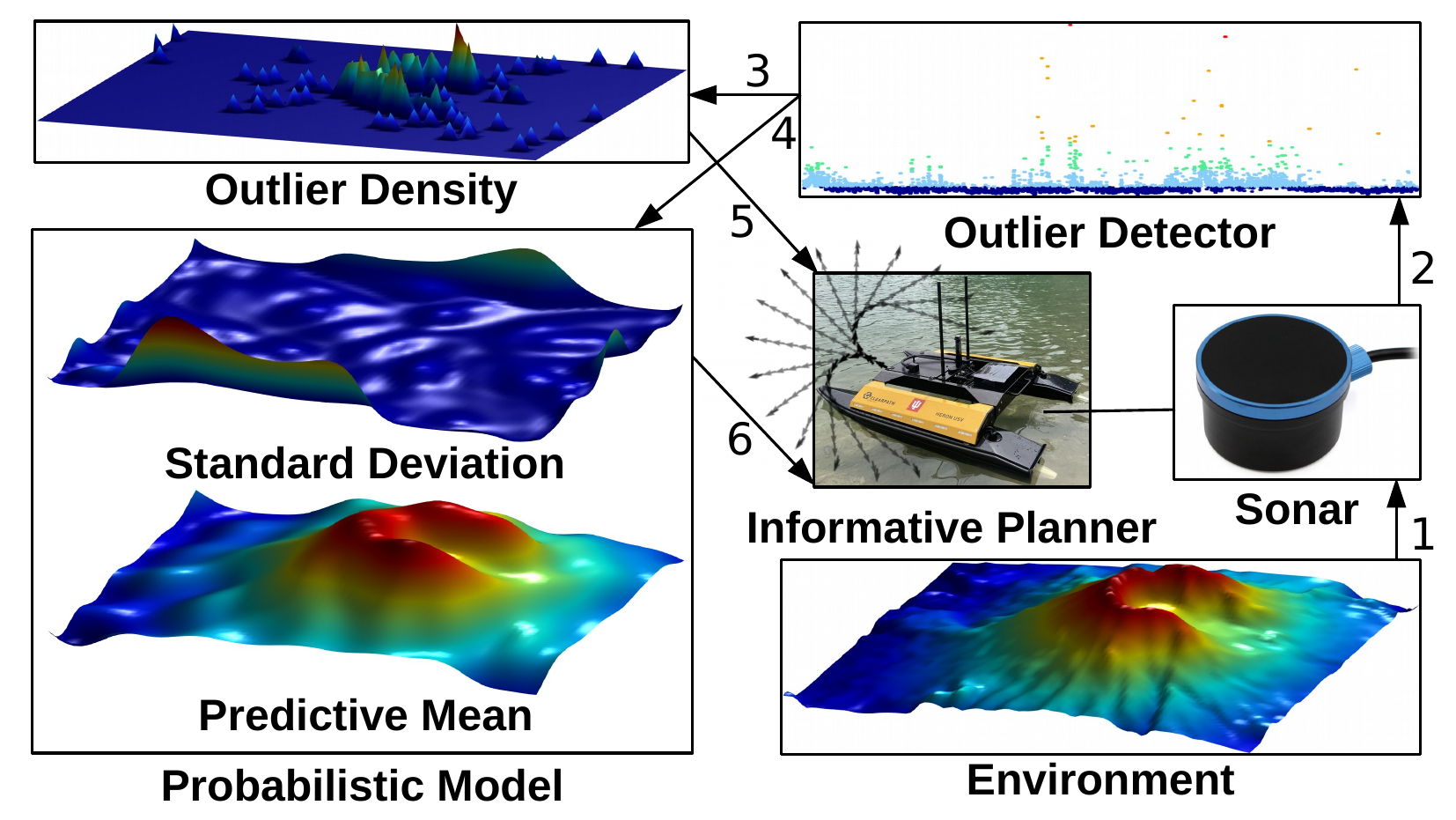}
  \caption{\textbf{System diagram}. (1) The Autonomous Surface Vehicle (ASV) collects depth measurements via the down-facing sonar. (2) An outlier detector keeps track of the outlier density across spatial locations (3) and passes the filtered data to the probabilistic model (4). The outlier density (5) and the predictive standard deviation (6) are fed to the multi-objective informative planner to guide the robot to collect the next batch of data.}%
  \label{fig:system}
\end{figure}

\cref{fig:system} illustrates the system with different modules and their relationship at a high level. The outlier detector takes raw sensing data, filters out the outliers, and outputs an outlier matrix indicating the density of outliers at each location (\cref{sec:copod}). The clean data is fed into the GPR model which produces the predictive mean and standard deviation at each location (\cref{sec:gpr}). The standard deviation matrix and the outlier matrix together form the multi-objective reward for the Pareto Monte Carlo tree search (\cref{sec:pareto}). Finally, the robot collects samples along the optimal trajectory given by the planner.

\subsection{Copula-Based Outlier Detection}%
\label{sec:copod}
COPula-based Outlier Detector (COPOD) is one of the top performing outlier detector that is efficient and tuning-free~\cite{li2020copod}. It is available in the open-source PyOD library~\cite{zhao2019pyod} and can be readily used in our problem. Firstly, COPOD fits empirical left tail cumulative distribution functions (CDFs) $\mathrm{F}^{l}_{1}(y)\dots,\mathrm{F}^{l}_{O}(y)$ using
\begin{equation}
  \mathrm{F}^{l}_{o}(y)=\mathbb{P}((-\infty,y])=\frac{1}{N}\sum_{n=1}^{N}\mathbb{I}(y_{n,o}\leq{y}),
\end{equation}
and empirical right tail CDFs $\mathrm{F}^{r}_{1}(y)\dots,\mathrm{F}^{r}_{O}(y)$ by replacing $y$ with $-y$, where $O$ is the observation dimension. The skewness vector $\mathbf{b}=[b_{1},\dots,b_{O}]$ is also computed using the standard estimation formula
\begin{equation}
  b_{o}=\frac{\frac{1}{N}\sum_{n=1}^{N}{(y_{n,o}-\overline{y_{o}})}^{3}}{\sqrt{\frac{1}{N-1}\sum_{n=1}^{N}{(y_{n,o}-\overline{y_{o}})}^{2}}^{3}},
\end{equation}
where the overline $\overline{y_{o}}$ denotes the mean. Then, we apply $F^{l}_{o}(y)$ and $F^{r}_{o}(y)$ on each sample to get the empirical copula observations $l_{n,o}$and $r_{n,o}$, respectively. The skewness corrected empirical copula observations are given by
\begin{equation}
  s_{n,o}=\begin{cases}
    l_{n,o} & \text{ if } b_{o}<0,\\
    r_{n,o} & \text{ otherwise}.
  \end{cases}
\end{equation}
Lastly, for each dimension, we compute the probability of observing a point at least as extreme as each $y_{n,o}$. The outlier score is the maximum negative log probability given by the left-tail, right-tail, and skewness-corrected empirical copula. Intuitively, outliers should occur less frequently, which means that the tail probabilities should be small, or equivalently, its negative log probabilities should be large.
We refer the reader to~\cite{li2020copod} for more details.

\subsection{Gaussian Process Regression}%
\label{sec:gpr}
A Gaussian process (GP) is a collection of random variables, any finite number of which have a joint Gaussian distribution~\cite{rasmussen2005mit}. We place a Gaussian process prior over the function
\begin{equation}
  f(\mathbf{x})\sim\mathcal{GP}(m(\mathbf{x}),k(\mathbf{x},\mathbf{x}'))
\end{equation}
which is specified by a mean function $m(\mathbf{x})$ and a covariance function $k(\mathbf{x},\mathbf{x}')$ (a.k.a. kernel). Popular choices of kernel functions include Mat\'ern and Gaussian kernels. In this paper, we use a zero mean and the anisotropic Gaussian kernel
\begin{equation}
  k(\mathbf{x},\mathbf{x}')=\alpha^{2}\exp{\left(-\frac{1}{2}\sum_{d=1}^{D}\frac{(x_{d}-x_{d}')^2}{\ell_{d}^{2}}\right)}.
  \label{eq:kernel}
\end{equation}
The amplitude $\alpha^{2}$ controls the variance of the random functions and $\ell_{d}$ defines the lengthscale, which informally can be thought of as the ``radius of a neighborhood'' in the $d$-th dimension. GP Regression (GPR) augments the model by assuming observations are corrupted by an additive Gaussian white noise
\begin{equation}
  p(y|\mathbf{x})=\mathcal{N}(y|f(\mathbf{x}),\sigma^2).
  \label{eq:likelihood}
\end{equation}
We will collect the parameters into $\bm{\theta}\triangleq\{\alpha,\ell_{d},\sigma\}$ which are termed as  \textit{hyperparameters}.

According to the definition of GP and the likelihood in \cref{eq:likelihood}, the joint distribution of the observations $\mathbf{y}$ and a set of latent function values $\mathbf{f}_{\star}$ at arbitrary test input locations $\mathbf{X}_{\star}$ is a multivariate Gaussian distribution
\begin{equation}
  \begin{bmatrix}
    \mathbf{y}\\
    \mathbf{f}_{\star}
  \end{bmatrix}\sim\mathcal{N}
  \left(
    \mathbf{0},
    \begin{bmatrix}
      \mathbf{K}_{y} & \mathbf{K}_{\star}\\
      \mathbf{K}_{\star}^{\T} & \mathbf{K}_{\star\star}
    \end{bmatrix}
  \right),
\end{equation}
where $\mathbf{K}_{y}=\mathbf{K}_{\mathbf{x}}+\sigma^2\mathbf{I}$, $\mathbf{K}_{\mathbf{x}}$ is the covariance matrix given by the covariance function evaluated at each pair of observations, $\mathbf{K}_{\star}$ is the covariance matrix between observations and the test function values $\mathbf{f}_{\star}$, and $\mathbf{K}_{\star\star}$ is the covariance matrix of the test function values.
The predictive distribution for GPR is a conditional Gaussian distribution
\begin{align}
  p(\mathbf{f}_{\star}\rvert\mathbf{y})&=\mathcal{N}(\mathbf{f}_{\star}\rvert{\bm{\mu}},\bm{\Sigma}), \text{ where}\label{eq:predictive_distribution}\\
  \bm{\mu}&=\mathbf{K}_{\star}^{\T}\mathbf{K}_{y}^{-1}\mathbf{y},\label{eq:predictive_mean}\\
  \mathbf{V}&=\mathbf{K}_{\star\star}-\mathbf{K}_{\star}^{\T}\mathbf{K}_{y}^{-1}\mathbf{K}_{\star}\label{eq:predictive_covariance}.
\end{align}

Learning in Gaussian process regression refers to determining appropriate values of the hyperparameters, and the common approach is to maximize the log marginal likelihood of the observations
\begin{equation}
  \ln{p(\mathbf{y}|\bm{\theta})}=-\frac{1}{2}\left(\mathbf{y}^{\T}\mathbf{K}_{y}^{-1}\mathbf{y}+\ln{\det(\mathbf{K}_{y})}+N\ln(2\pi)\right),
  \label{eq:log_marginal_likelihood}
\end{equation}
where $\det(\cdot)$ denotes the matrix determinant.

\subsection{Pareto Monte Carlo Tree Search}%
\label{sec:pareto}
We treat the informative planning problem as a sequential decision making problem and approximate the solution using Monte Carlo tree search (MCTS)~\cite{browne2012survey}. A node in the tree contains a state/pose and some statistics. The MCTS iterates in four steps: \emph{node selection, expansion, simulation}, and \emph{back-propagation}. We now describe the reward function and each step of the MCTS in detail.

\begin{figure}[thbp]%
  \centering%
  \subfloat[]{%
    \resizebox{0.5\linewidth}{!}{
      \begin{tikzpicture}%
        \node at(0.0,0.0){%
        \includegraphics[width=\linewidth,height=0.8\linewidth]{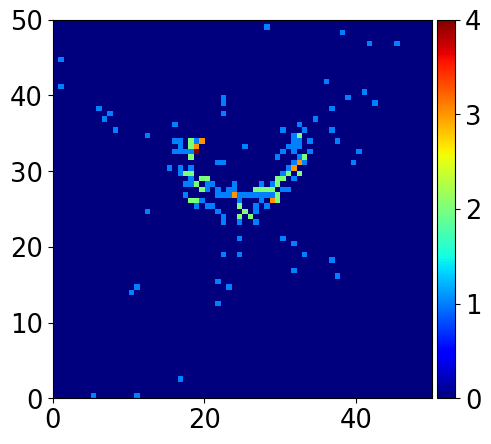}};%
      \end{tikzpicture}}}%
  \subfloat[]{%
    \resizebox{0.5\linewidth}{!}{
      \begin{tikzpicture}%
        \node at(0.0,0.0){%
        \includegraphics[width=\linewidth,height=0.8\linewidth]{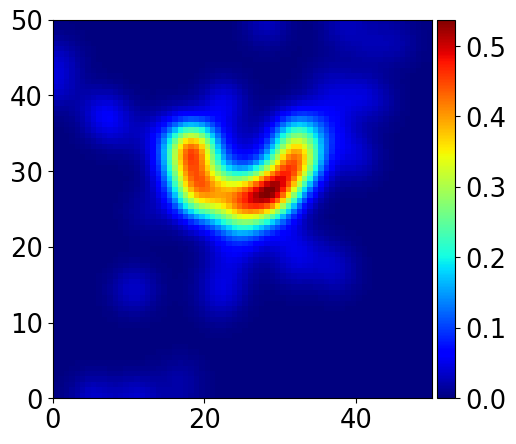}};%
      \end{tikzpicture}}}%
  \caption{\textbf{(a) A matrix storing the outlier count} at each spatial location. \textbf{(b) A smoothed version of (a)} using a Gaussian kernel. The smoothed outlier occurance matrix serves as one of the reward map in Pareto Monte Carlo tree search to revisiting sampling locations with outliers.}%
  \label{fig:outliers}%
\end{figure}

\noindent\textbf{Reward Functions}.
The reward of an action should be the amount of the ``informativeness'' of the collected data by taking this action. A natural choice for measuring informativeness is the mutual information between the visited locations and the remainder of the space~\cite{krause2008near}. However, calculating entropy involves the determinant computation of the predictive covariance matrix of the GPR in \cref{eq:predictive_covariance}, which is computationally costly. A cheaper approximation is the reduction in the trace of the predictive covariance matrix, namely, variance reduction. Observing that the predictive uncertainty at a location becomes very small after sampling at that location, we further simplify the maximization of variance reduction to simply maximizing the sum of predictive standard deviation along the sampling trajectory. To encourage re-visiting the locations where outliers were detected, another reward is the number of outlier occurrence. For better planning efficiency, we pre-compute the predictive standard deviation and the number of outliers on a \emph{query grid} representing all discretized sampling locations before the tree search process. When evaluating the reward function values, we simply access the corresponding matrix elements. To better guide the robot, we also filter the outlier occurrence matrix via a Gaussian kernel. \cref{fig:outliers} shows the outlier occurrence matrix and its smoothed version.

\noindent\textbf{Selection.}
As a best-first search algorithm, MCTS selects nodes that are expected to have a high reward but at the same time try other nodes sufficiently to avoid greedy choices. We trade-off \emph{exploration} and \emph{exploitation} using the tree policy. A well known tree policy is the upper confidence bound (UCB)~\cite{auer2002using,kocsis2006bandit}. We recursively select child nodes with the highest UCB value until a node with unexpanded children is found. For node $j$, the UCB value adapted to our problem is defined by
\begin{equation}
  \operatorname{UCB}_{j}=\overline{f_{i}}_{j}+C\sqrt{\frac{2 \log N_{p}}{N_{j}}},
  \label{eq:ucb}
\end{equation}
where $N_{p}$ is the number of times that the parent of node $j$ has been selected and $N_{j}$ is the number of times that node $j$ has been selected. The average reward $\overline{f_{i}}_{j}$ encourages the selection of nodes that currently look the most promising and thus stimulates exploitation. The value of the second term is large when $N_{j}$ is small, which encourages exploring node $i$ if it is not selected enough times. Constant $C$ balances between the exploration and the exploitation which should be set to a similar scale as the rewards.

For vector reward given by the multiple reward functions, we use the Pareto variant of UCB proposed in~\cite{chen2019pareto}:
\begin{equation}
  \operatorname{PUCB}_{j}=\overline{\mathbf{f}_{i}}_{j}+C\sqrt{\frac{4\log N_{p}+\ln{D_{r}}}{2 * N_{j}}},
  \label{eq:pucb}
\end{equation}
where $\overline{\mathbf{f}_{i}}_{j}$ is the average reward vector and $D_{r}$ is the dimensionality of this vector. When selecting the best child node, we first compute the Pareto optimal set from the Pareto UCB vectors of all child nodes. The vectors in the Pareto optimal set cannot be improved for any objective without hurting other objectives. We then randomly select a child node from this set because the Pareto optimal solutions are considered equally optimal if no preference information is given. We refer the reader to~\cite{chen2019pareto} for further details of how Pareto MCTS balances multiple objectives. When the UCB tree policy is applied to a MCTS, the algorithm is commonly referred to as UCT.
Similarly, we use \emph{PUCT} to represent Pareto UCB applied to MCTS.

\noindent\textbf{Expansion}.
We randomly select an action from the selected node and delete this action from the list of available actions. We then take this action and calculate the corresponding reward. A new node is created based on the new state and reward. We append this new node to the children list of the selected node.

\noindent\textbf{Simulation}.
The goal of this step is to estimate the expected reward of the action of the newly expanded node by executing a default/rollout policy. Here we define the rollout policy to be ``moving forward'' for some steps. This policy is designed to estimate the expected reward along the direction of executing an action.

\noindent\textbf{Back-propagation}.
We add the average reward received in simulation to each node along the selection path and increase the number of visits of these nodes. These four steps are repeated until the computational budget for the robot has reached. We recursively select the child node with the highest number of visits to get the final informative action sequence.

{\algrenewcommand\textproc{}
  \begin{algorithm}[tp]\small
    \setstretch{1.3}
    \caption{\textbf{The Proposed Framework}}\label{alg:overview}
    \begin{algorithmic}[1] 
      \State Collect the initial training data 
      \State Initialize and optimize \texttt{gpr} with the initial data
      \State Initialize the outlier matrix \texttt{outlier} to be zeros
      \While {has sampling budget}
      \State \texttt{mean,std=gpr.predict(x\_test)}
      \State \texttt{trajectory=PUCT(pose,std,outlier)}
      \State Follow \texttt{trajectory} and sample new data
      \State Train the outlier detector on all collected data
      \State Filter out the outliers in the new data
      \State \texttt{gpr} appends the new data and optimize for several iterations
      \EndWhile
    \end{algorithmic} 
  \end{algorithm}
}

\begin{figure}[thbp]%
  \centering%
  \subfloat[]{%
    \resizebox{0.5\linewidth}{!}{
      \begin{tikzpicture}%
        \node at(0.0,0.0){%
        \includegraphics[width=\linewidth,height=0.7\linewidth,trim={38 25 0 0},clip]{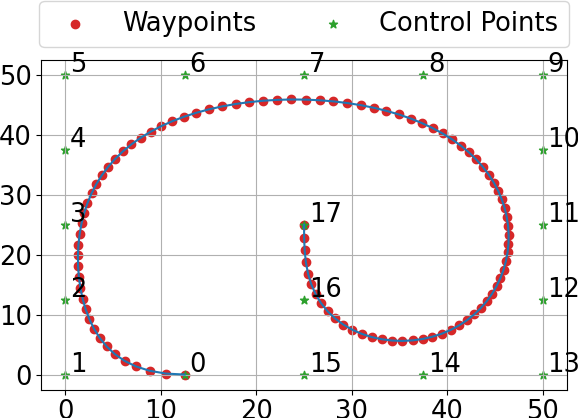}};%
      \end{tikzpicture}}}%
  \subfloat[]{%
    \resizebox{0.5\linewidth}{!}{
      \begin{tikzpicture}%
        \node at(0.0,0.0){%
        \includegraphics[width=\linewidth,height=0.7\linewidth]{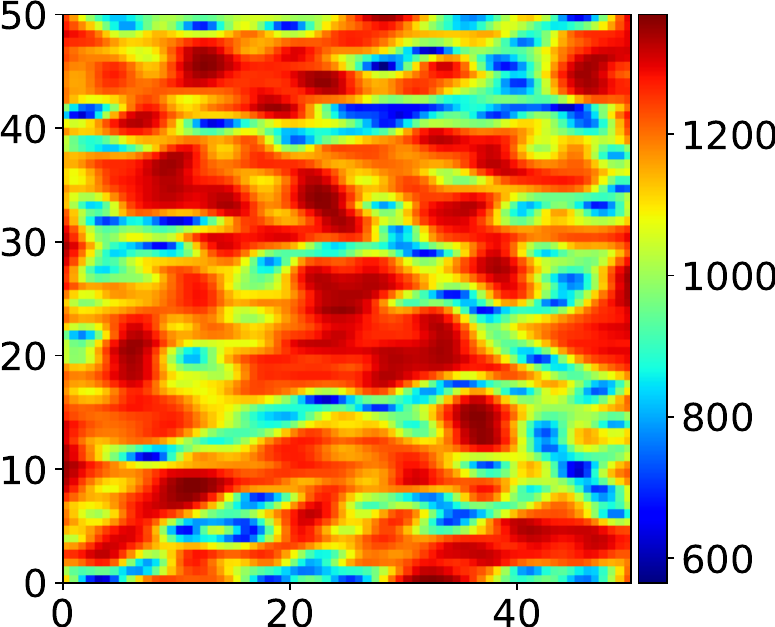}};%
      \end{tikzpicture}}}%
  \caption{(a) \textbf{The Bezier curve} to generate \emph{a priori} sampling path for collecting initial training data for the initial optimization of the GPR and computing data pre/post-processing statistics. (b) \textbf{Final predictive standard deviation} of UCT-NONE. }%
  \label{fig:bezier_std}%
\end{figure}

\begin{figure*}[tp]%
  \centering%
  \subfloat[$\rho=0.05$]{%
    \resizebox{0.43\linewidth}{!}{
      \begin{tikzpicture}%
        \node at(0.0,0.0){%
        \includegraphics[width=\linewidth]{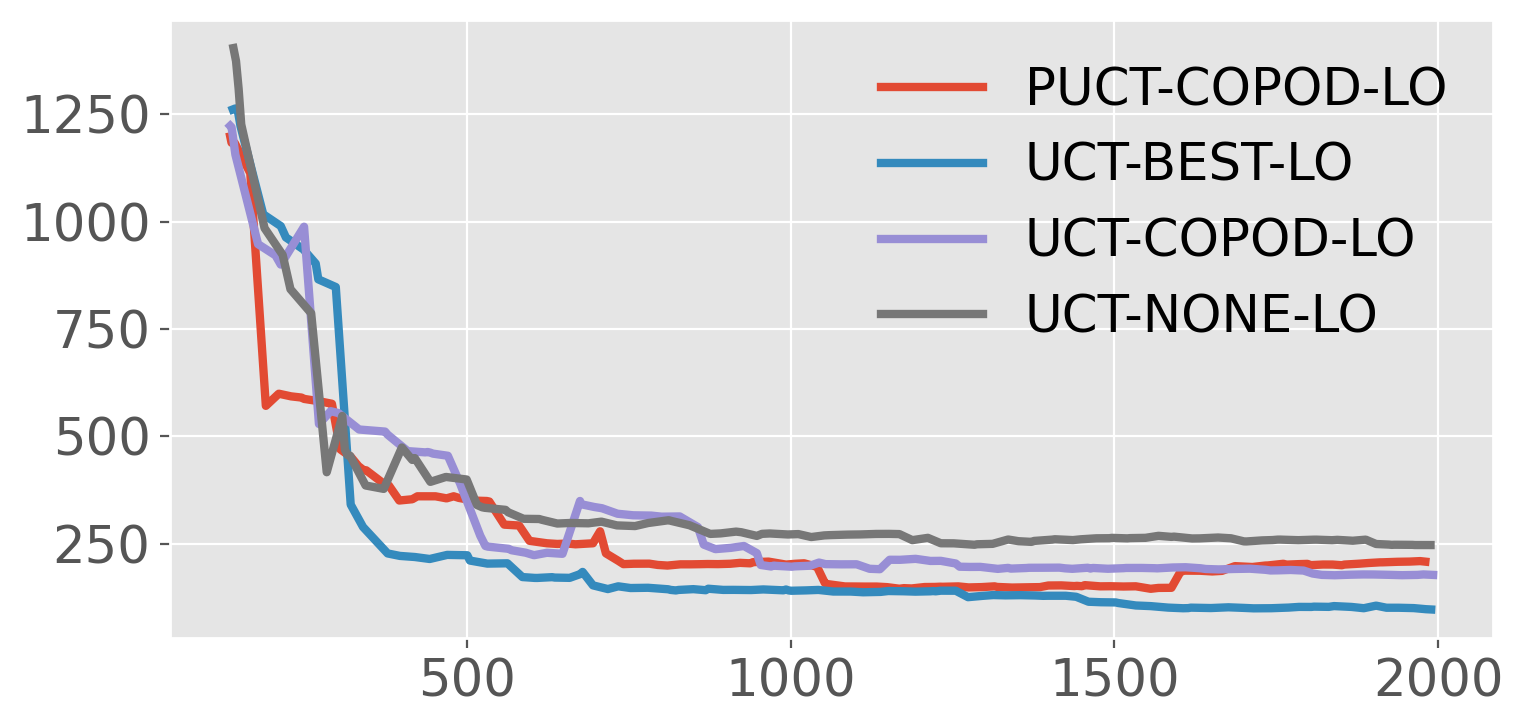}};%
      \end{tikzpicture}%
    }%
  }%
  \subfloat[$\rho=0.10$]{%
    \resizebox{0.43\linewidth}{!}{
      \begin{tikzpicture}%
        \node at(0.0,0.0){%
        \includegraphics[width=\linewidth]{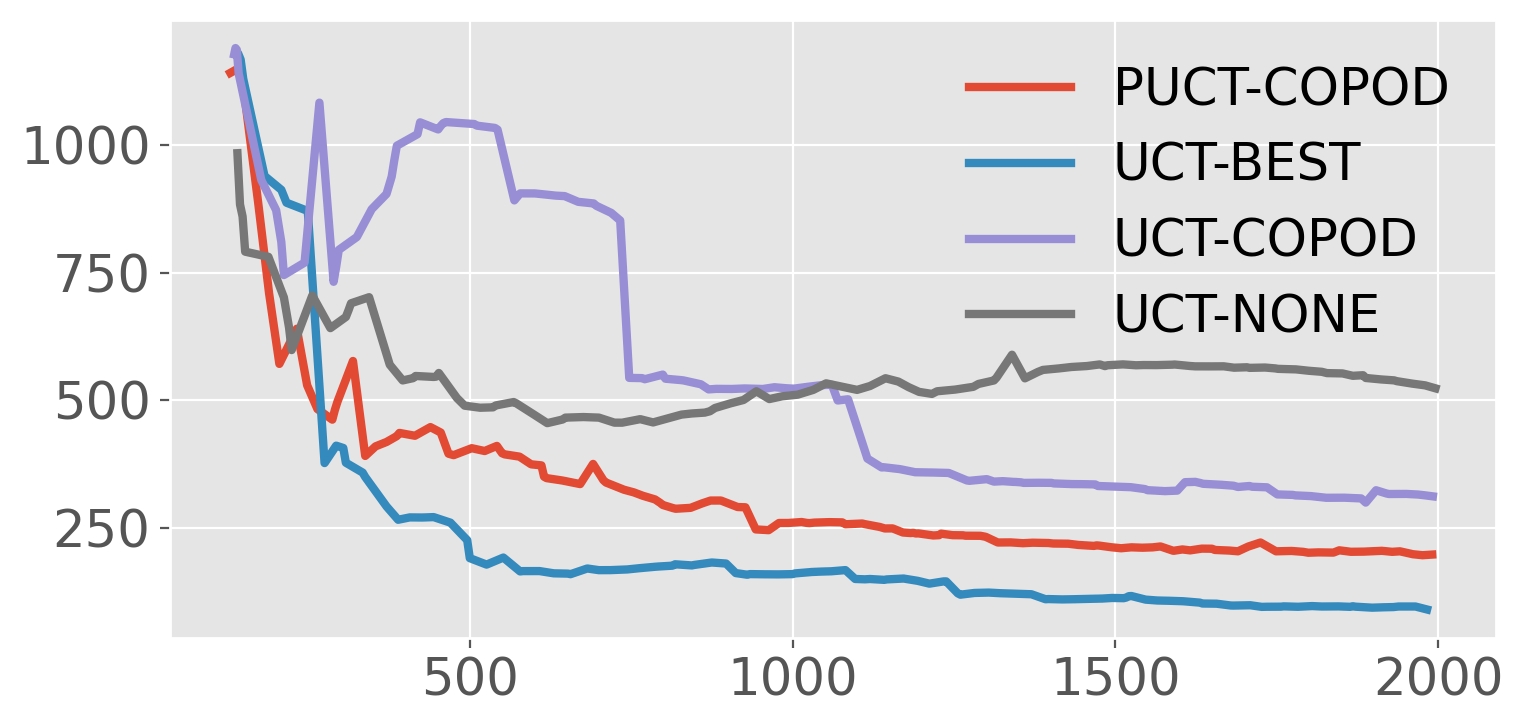}};%
      \end{tikzpicture}%
    }%
  }\\%
  \subfloat[$\rho=0.15$]{%
    \resizebox{0.43\linewidth}{!}{
      \begin{tikzpicture}%
        \node at(0.0,0.0){%
        \includegraphics[width=\linewidth]{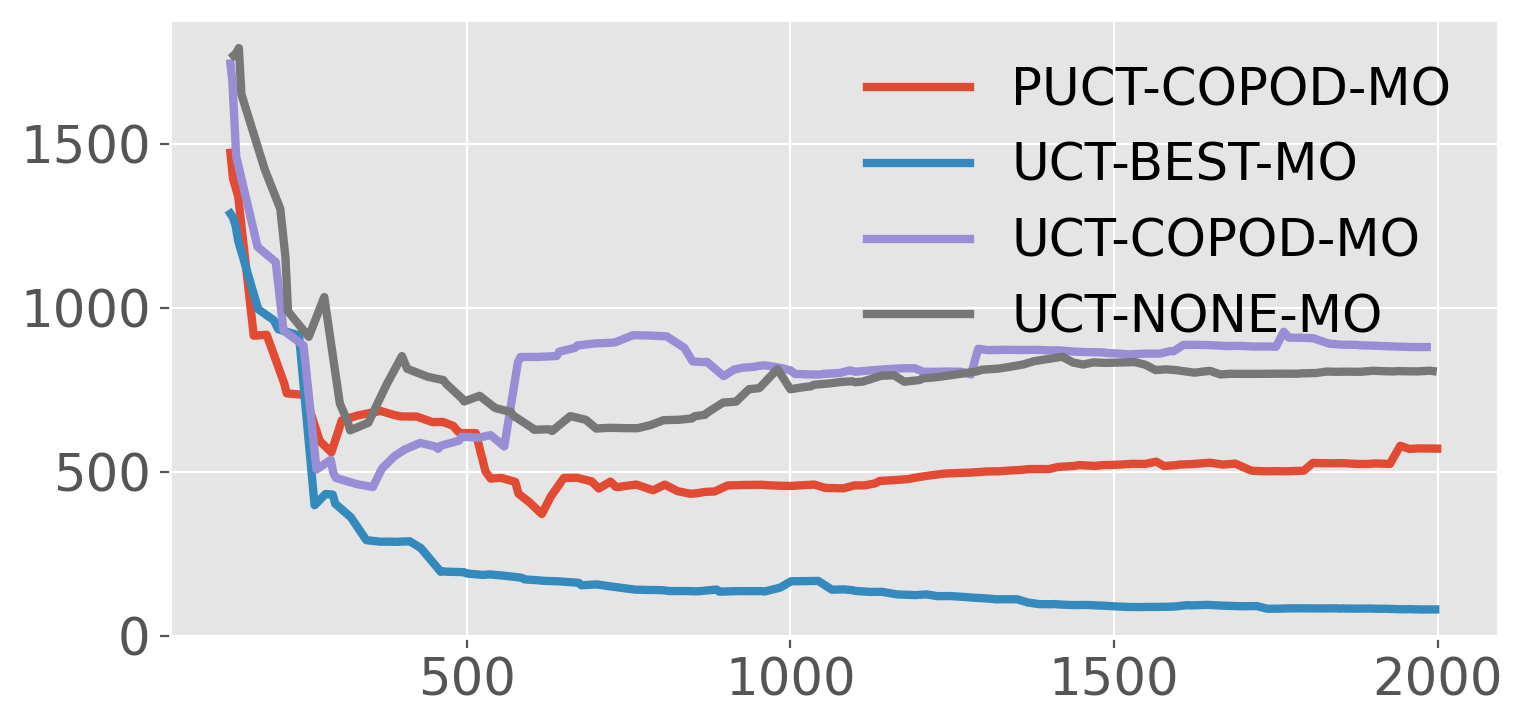}};%
      \end{tikzpicture}%
    }%
  }%
  \subfloat[$\rho=0.10$]{%
    \resizebox{0.43\linewidth}{!}{
      \begin{tikzpicture}%
        \node at(0.0,0.0){%
        \includegraphics[width=\linewidth]{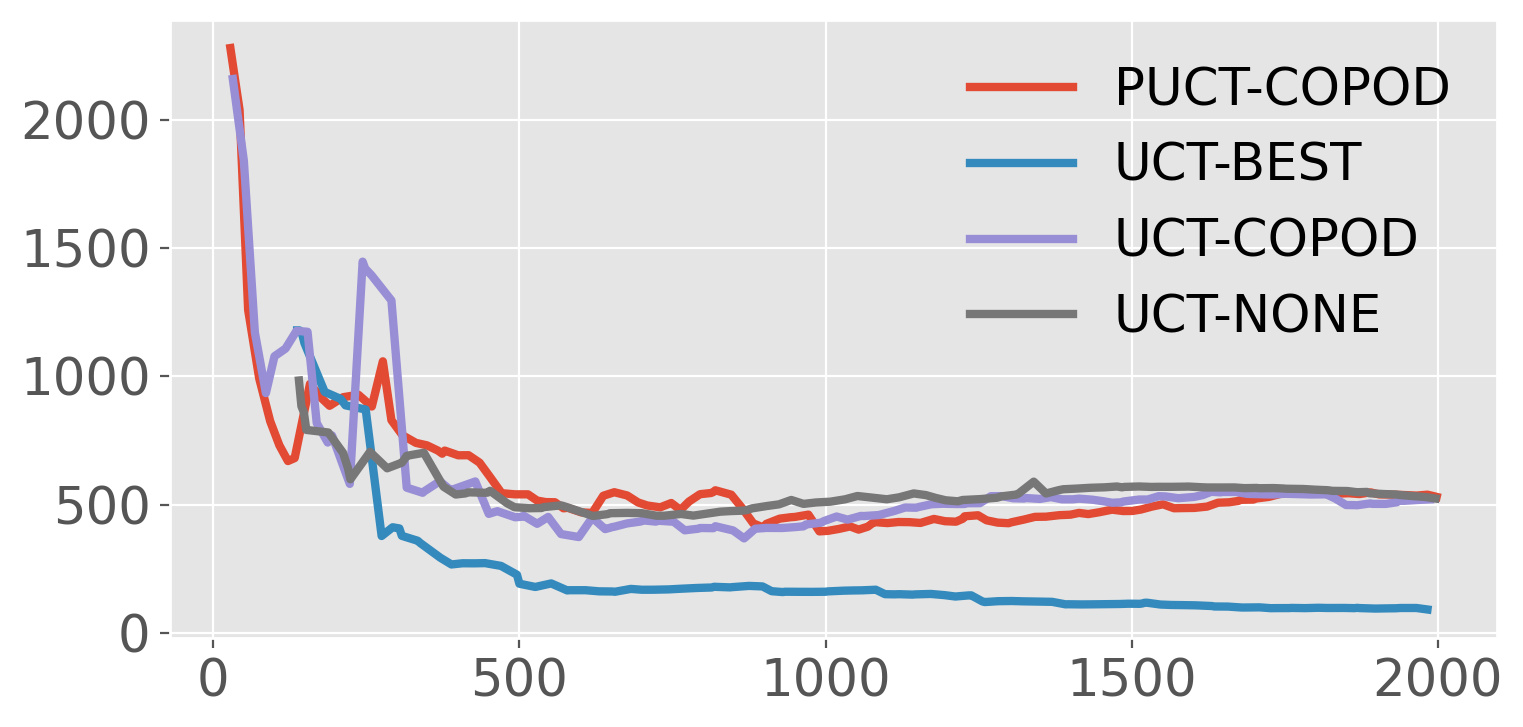}};%
      \end{tikzpicture}}}%
  \caption{\textbf{Root mean squared errors} of the proposed PUCT-COPOD and other three baselines. UCT-BEST filters outliers with the ground-truth labels, so it serves as a best-case baseline; UCT-NONE, which is the worst case baseline, does not filter outliers at all. UCT-COPOD directly applies the outlier detector to the sensing data stream. Parameter $\rho$ controls the number of outliers: higher value indicates more outliers (MO) while lower value means less outliers (LO). In (a)-(c), the outlier detector only filters out the outliers in the newly acquired batch without changing the historical data. In (d), the outlier detector re-examine \emph{all} the historical data at each decision epoch. (a) Performance difference of all the methods is insignificant when the number of outliers is very small. (b) PUCT-COPOD and UCT-COPOD are both in between the best-case baseline and the worst-case baseline after collecting $1000$ samples while PUCT-COPOD performs better. At the early stage, UCT-COPOD is even worse than UCT-NONE because it filters out some informative samples. (c) When the number of outliers increases, UCT-COPOD loses advantage over UCT-NONE. The performance of PUCT-COPOD is also affected but its still better than UCT-COPOD and UCT-NONE. (d) Both PUCT-COPOD and UCT-COPOD are ineffective if the detector filters outliers in all historical data.}%
  \label{fig:rmse}%
\end{figure*}

\subsection{Overall Framework}%
\label{sec:framework}
Algorithm\,\ref{alg:overview} summarized main steps of the proposed framework. First, we collect the \emph{pilot} data following a path that does not depend on the informative planning framework. For example, we will use Bezier curve in \cref{fig:bezier_std}\,(a). The pilot data is used for an initial optimization of the GPR hyperparameters through \cref{eq:log_marginal_likelihood} and computing the statistics for pre/post-processing of the data. In a decision epoch, the robot searches for an optimal trajectory using PUCT based on the predictive standard deviation given by the GPR and the outlier matrix given by the outlier detector. The robot then collects samples along the trajectory. Finally, we detect outliers in this batch of new samples, get rid of them, feed the clean data to the GPR, and optimize the hyperparameters for several iterations.

\begin{figure*}[tp]%
  \centering%
  \subfloat[Ground truth\label{fig:gt}]{%
    \resizebox{0.33\linewidth}{!}{
      \begin{tikzpicture}%
        \node at(0.0,0.0){%
        \includegraphics[width=\linewidth]{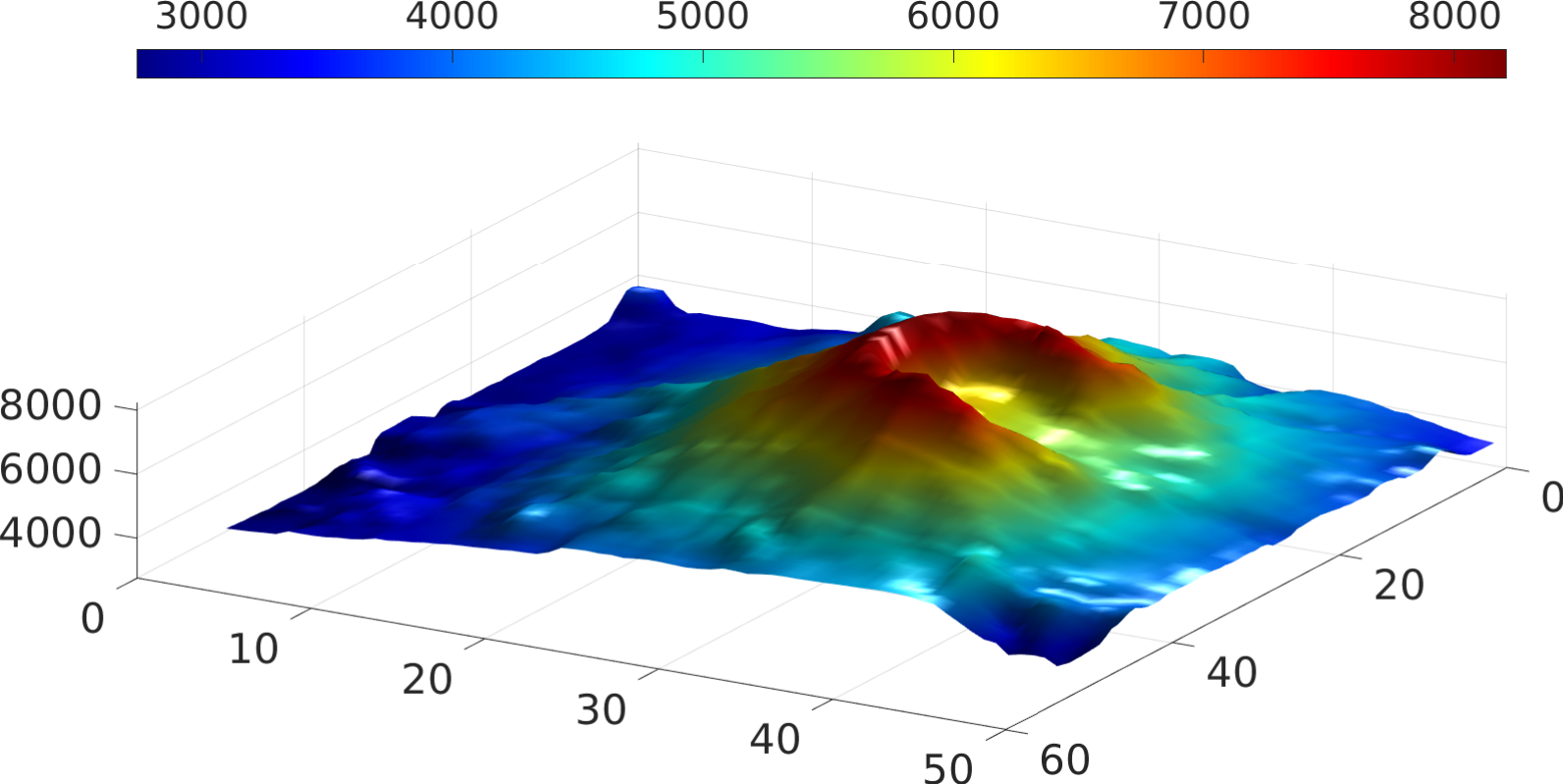}};%
      \end{tikzpicture}}}%
  \subfloat[PUCT-COPOD\label{fig:puct_copod}]{%
    \resizebox{0.33\linewidth}{!}{
      \begin{tikzpicture}%
        \node at(0.0,0.0){%
        \includegraphics[width=\linewidth]{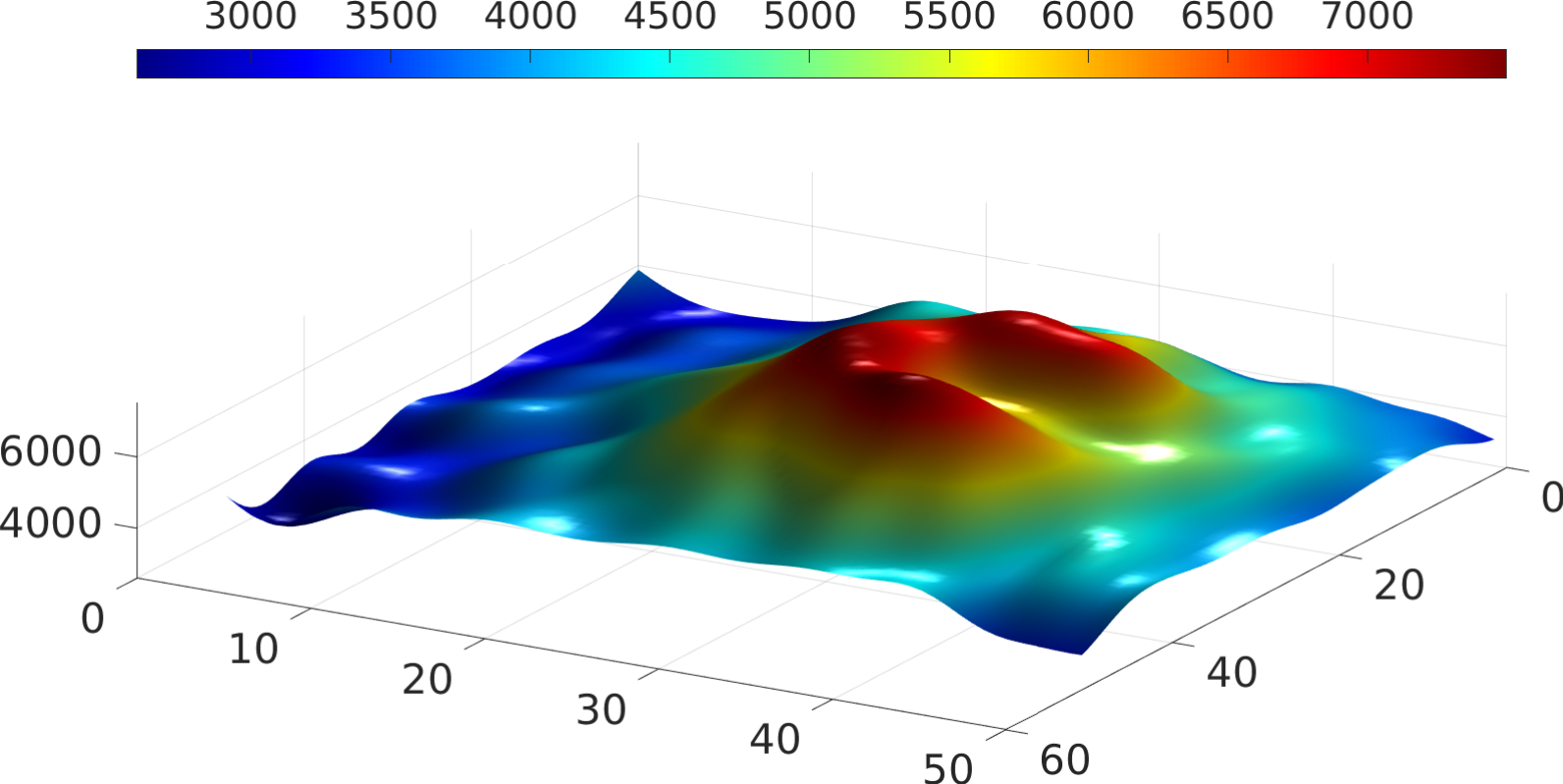}};%
      \end{tikzpicture}}}%
  \subfloat[UCT-COPOD\label{fig:uct_copod}]{%
    \resizebox{0.33\linewidth}{!}{
      \begin{tikzpicture}%
        \node at(0.0,0.0){%
        \includegraphics[width=\linewidth]{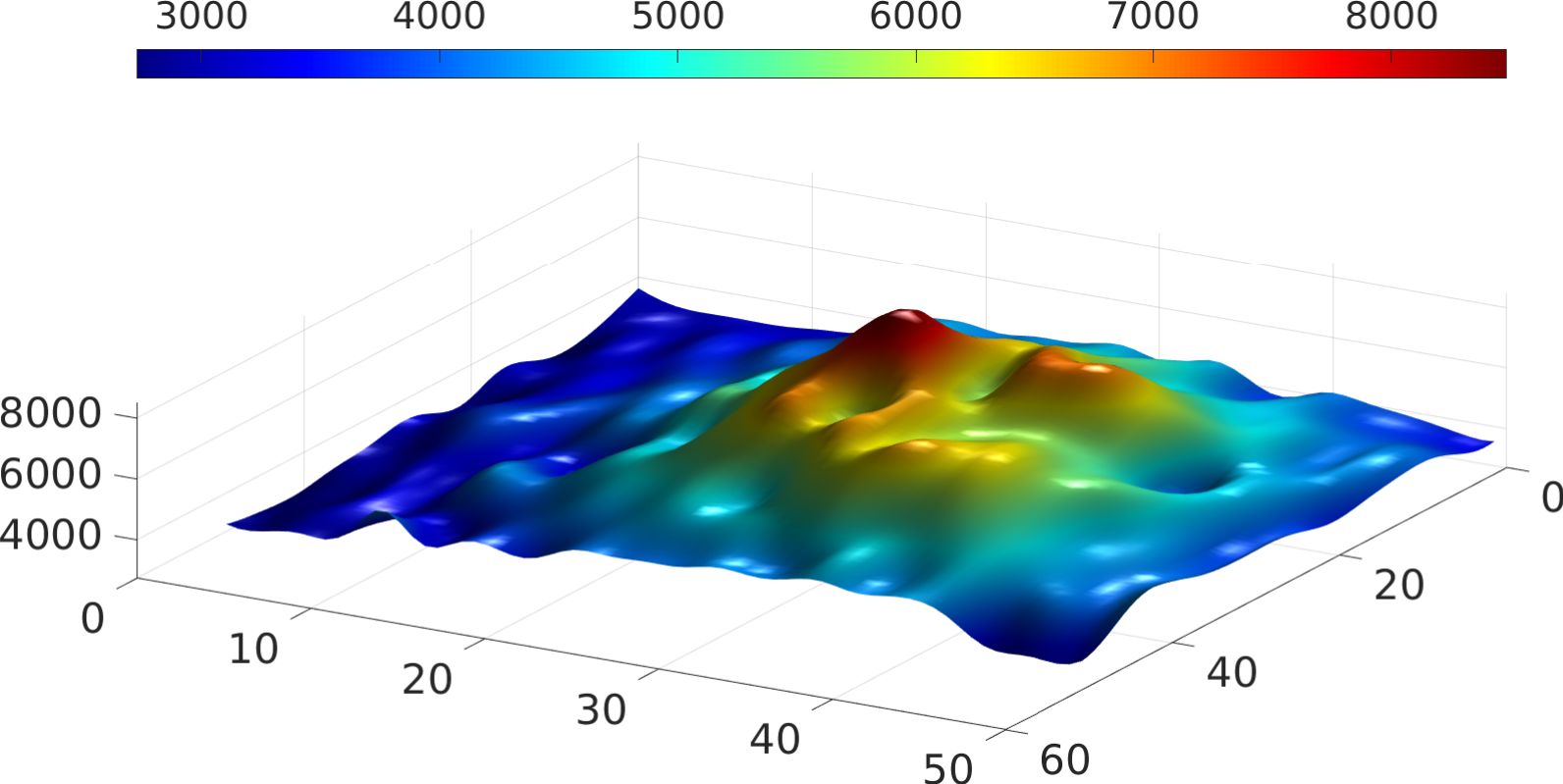}};%
      \end{tikzpicture}}}%
\caption{\textbf{Final predictive means} of PUCT-COPOD and UCT-COPOD.}%
  \label{fig:prediction}%
\end{figure*}

\begin{figure}[thbp]%
  \centering%
  \subfloat[UCT-COPOD]{%
    \resizebox{0.5\linewidth}{!}{
      \begin{tikzpicture}%
        \node at(0.0,0.0){%
        \includegraphics[width=\linewidth]{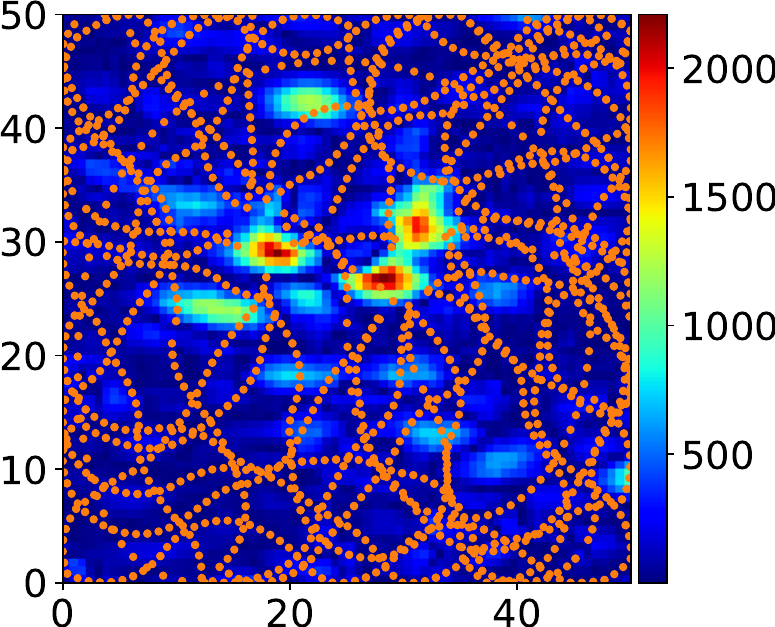}};%
      \end{tikzpicture}%
    }%
  }%
  \subfloat[PUCT-COPOD]{%
    \resizebox{0.5\linewidth}{!}{
      \begin{tikzpicture}%
        \node at(0.0,0.0){%
        \includegraphics[width=\linewidth]{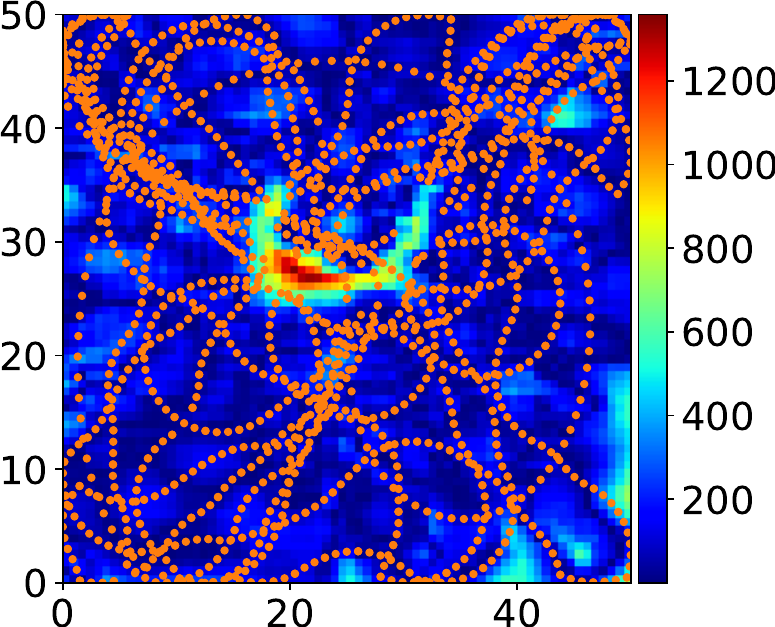}};%
      \end{tikzpicture}%
    }%
  }%
  \caption{\textbf{The absolute error maps and training samples after cleaning the outliers} of (a) UCT-COPOD and (b) PUCT-COPOD, respectively. Many informative samples at the high-variability region are falsely filtered out in (a), leading to some high-error ``holes''. As a comparison, (b) shows that the proposed method can mitigate this effect to some extent.}%
  \label{fig:samples}%
\end{figure}

\section{Experimental Results}%
\label{sec:experiments}
We would like to answer the following questions.
\begin{itemize}
  \item How do outliers affect an informative planning system if they are not filtered out?
    To answer this question, we use the standard GPR and UCT without any outlier detector as the worst-case baseline (UCT-NONE).
  \item What is the ideal performance if we have a perfect outlier detector? We use the ground-truth outlier labels as the prediction of a perfect outlier detector. We refer to this best-case competitor as UCT-BEST.
  \item What if we simply apply an off-the-shelf detector to deal with the outliers? In this method, we use COPOD to get rid of outliers (UCT-COPOD). 
  \item Is the proposed method able to mitigate the negative effect of outliers? We denote our proposed method as PUCT-COPOD.
\end{itemize}

Considering that the second question requires the ground-truth outlier labels and to better understand the results, we inject some ``spike'' outliers to the sensing data which are similar to those in \cref{fig:pull_figure}\,(a). The robot collects a batch of data along the planned trajectory in each decision epoch. We randomly select $\rho$ portion of the batch for outlier injection. Specifically, we first sample a random amplitude uniformly from the range $[1, 2]$ with a random sign. Then, the amplitude is multiplied with the data range of the batch computed by the $0.05$ and $0.95$ quantiles. We use the elevation map of the Mount Saint Helens, shown in \cref{fig:gt}, as the environment. The planar robot follows the Dubins car dynamics:
$
  \begin{bmatrix}
    \dot{x}_1,
    \dot{x}_2,
    \dot{\theta}
    \end{bmatrix}=\begin{bmatrix}
    v\cos{\theta},
    v\sin{\theta},
    u
  \end{bmatrix},u\in\mathcal{U},
  \label{eq:motion_model}
$
where $\mathbf{x}=[x_1, x_2]^\intercal\in\mathbb{R}^2$ and $\theta\in[0,2\pi)$ are the position and orientation of the vehicle, $u$ is the control input that represents the robot's steering angle. We set the linear velocity $v$ to be a constant and only control the steering angle $u$. The possible steering angles $[-0.15, 0.15]$ (in radians) are discretized into $5$ choices.

The GPR is initialized using $100$ training samples generated by a Bezier curve shown in \cref{fig:bezier_std}\,(a) and optimize the hyperparameters for $500$ iterations. This \emph{a priori} Bezier sampling path can be applied to any environment for collecting the \emph{pilot} data for the initial GPR optimization and computing data pre/post-processing statistics. The path circles around the whole environment to help us get a sense of the workspace extent and a rough estimate of the observation range. Using the statistics computed from the initial training data, all inputs $\mathbf{X}$ will be scaled to the range of $[-1, 1]$ and all observations $\mathbf{y}$ will be standardized to have zero mean and unit variance. We use $500$ tree search iterations and $5$ rollout iterations. The sampling budget is set to $2000$ samples.

\cref{fig:rmse} shows the root mean squared error (RMSE) of the four methods versus the number of training samples. When the number of outliers is very small ($\rho=0.05$ in \cref{fig:rmse}\,(a)), the difference among all the methods is negligible. After increasing $\rho$ to $0.1$, as expected, the error of UCT-BEST drops drastically and reaches the lowest error. After the decrease at the beginning, the error of UCT-None is actually increasing. The inclusion of outliers not only makes the prediction deteriorated but also messes up the standard deviation (see \cref{fig:bezier_std}\,(b)), which further affects the downstream informative planning. The result of UCT-COPOD is interesting --- although the final error is lower than that of UCT-NONE, it is the worst at the beginning. The reason for this can be seen from \cref{fig:samples}\,(a). Many informative samples at the high-variability region are falsely filtered out, leading to some high-error ``holes''. As a comparison, \cref{fig:samples}\,(b) shows that the proposed method can mitigate this effect to some extent. This brings better RMSE curve to PUCT-COPOD in \cref{fig:rmse}\,(b). When there are more outliers ($\rho=0.15$), UCT-COPOD loses the advantage over UCT-NONE. The performance of PUCT-COPOD is also affected but its still better than UCT-COPOD and UCT-NONE. Finally, we also study the case where the detector filters outliers in all the historical data instead of the newly acquired batch. Both PUCT-COPOD and UCT-COPOD are ineffective in this case. One way to understand the result is that, when the robot has not sampled the volcanic area, the detector can easily detect extremely high or low values because the detector is trained on the historical data sampled from the flatland. However, after collecting samples from the volcanic area and retaining the detector, it is difficult for the detector to distinguish the aforementioned extreme values and the normal elevation measurements around the volcanic area during re-examination. In addition to the quantitative difference in the error reduction, the final predictions of the proposed method and the baseline are also qualitatively different. \cref{fig:prediction} shows that the prediction of PUCT-COPOD is closer to the ground-truth environment than that of UCT-COPOD.

\section{Conclusion}
\label{sec:conclusion}%
We present a framework to enable the robot to re-visit the locations where outliers were sampled besides optimizing the conventional informative planning objective. This is very different from existing informative planning approaches which mainly focus on planners that optimize over uncertainty or confidence based informativeness, which can be easily deteriorated by various sensing outliers, resulting in a misleading objective and sub-optimal sampling behaviors. We propose an approach to filter out the outliers from the sensing data stream using an off-the-shelf outlier detector. By designing a new planning objective with a Pareto variant of Monte Carlo tree search, our new framework allows the robot to collect more samples in the vicinity of the outliers and update the outlier detector to reduce the number of false alarms. Results show that the proposed framework performs much better than only applying the outlier detector.

\section{Acknowledgement}
\label{sec:acknowledgement}%
We gratefully acknowledge the support of  NSF grants with grant numbers 1906694, 4848921, and 4848922. 
We also thank the valuable comments from anonymous reviewers.  


\clearpage
\bibliographystyle{unsrtnat}
\balance
\bibliography{references}
\end{document}